\documentclass[sn-vancouver]{sn-jnl}

\jyear{2023}%

\usepackage{color}
\usepackage{makecell}
\usepackage{rotating}
\usepackage{pifont}
\newcommand{\cmark}{\textcolor{green}{\ding{51}}}%
\newcommand{\xmark}{\textcolor{red}{\ding{55}}}%

\theoremstyle{thmstyleone}%
\theoremstyle{thmstyletwo}%

\theoremstyle{thmstylethree}%

\newcommand{\custompara}[1]{\vspace*{1ex}\noindent\textbf{#1.}\hspace*{1ex}}

\begin{document}

\title[Formal Modelling for Multi-Robot Systems Under Uncertainty]{\vspace{-20pt}Formal Modelling for Multi-Robot Systems Under Uncertainty}


\author*[1]{\fnm{Charlie} \sur{Street}}\email{c.l.street@bham.ac.uk}

\author[1]{\fnm{Masoumeh} \sur{Mansouri}}\email{m.mansouri@bham.ac.uk}

\author[2]{\fnm{Bruno} \sur{Lacerda}}\email{bruno@robots.ox.ac.uk}

\affil[1]{\orgdiv{School of Computer Science}, \orgname{University of Birmingham}, \orgaddress{\country{UK}}}

\affil[2]{\orgdiv{Oxford Robotics Institute}, \orgname{University of Oxford}, \orgaddress{\country{UK}}}


\abstract{\textbf{Purpose of Review:} To effectively synthesise and analyse multi-robot behaviour, we require formal task-level models which accurately capture multi-robot execution. 
In this paper, we review modelling formalisms for multi-robot systems under uncertainty, and discuss how they can be used for planning, reinforcement learning, model checking, and simulation.\\
\textbf{Recent Findings:} Recent work has investigated models which more accurately capture multi-robot execution by considering different forms of uncertainty, such as temporal uncertainty and partial observability, and modelling the effects of robot interactions on action execution.
Other strands of work have presented approaches for reducing the size of multi-robot models to admit more efficient solution methods.
This can be achieved by decoupling the robots under independence assumptions, or reasoning over higher level macro actions.\\
\textbf{Summary:} Existing multi-robot models demonstrate a trade off between accurately capturing robot dependencies and uncertainty, and being small enough to tractably solve real world problems.
Therefore, future research should exploit realistic assumptions over multi-robot behaviour to develop smaller models which retain accurate representations of uncertainty and robot interactions; and exploit the structure of multi-robot problems, such as factored state spaces, to develop scalable solution methods.
}

\keywords{Multi-Robot Systems, Markov Models, Uncertainty\\ \\This version of the article has been accepted for publication, after peer review (when applicable)
but is not the Version of Record and does not reflect post-acceptance improvements, or any
corrections. The Version of Record is available online at: \url{http://dx.doi.org/10.1007/s43154-023-00104-0}.}


\maketitle

\section{Introduction}\label{sec:intro}

The demand for multi-robot systems (MRSs) is increasing, due to their performance, flexibility, and fault tolerance~\cite{iocchi2000reactivity,yan2013survey}.
Successful multi-robot deployments have been completed in a range of domains, such as fulfilment centres~\cite{ocadocfc}, fruit fields~\cite{khan2020incorporating}, and roads~\cite{robotaxifleet}.
For safe and robust multi-robot coordination in the real world, it is often desirable to consider \emph{formal models} of the MRS, which enable policy synthesis for well-defined objectives, as well as a formal analysis of such policies.
In this review paper, we consider formal models that capture the task-level behaviour of the MRS.
These model high-level capabilities such as navigation or manipulation, while abstracting the lower-level control required to implement these capabilities. 
Formal models are used alongside multi-robot planning~\cite{boutilier1996planning} and reinforcement learning (RL)~\cite{bucsoniu2010multi} techniques to synthesise robot behaviour, and alongside model checking~\cite{herd2015quantitative} and simulation~\cite{damas2004stochastic} techniques to evaluate task-level metrics of multi-robot performance.
However, the success of these techniques is limited by the model's accuracy, in particular its capacity to capture and predict execution-time multi-robot behaviour~\cite{baier2008principles}.
For example, if we plan on an inaccurate model, our expectations of robot behaviour during planning diverge from what is observed during execution, which can lead to inefficient execution-time behaviour, or robot failure in the worst case.

In this paper, we focus on modelling the \emph{stochasticity} of MRSs as, in any environment, robot behaviour is affected by the stochastic dynamics of the environment and the other robots.
For example, a mobile robot operating in an office may fail to navigate upon a door being closed unexpectedly; or it may be unable to dock at a charging station if another robot is charging for longer than expected.
We begin by introducing the types of uncertainty encountered by MRSs, including uncertainty over action outcomes~\cite{puterman1994mdp}, a robot's current state~\cite{kaelbling1998planning}, and the duration and start time of robot actions~\cite{boyan2001exact,street2022context}.
Next, we review modelling formalisms which capture these sources of uncertainty.
We then describe how formal multi-robot models have been used to support advances in the application of planning, RL, model checking, and simulation techniques to MRSs.

\section{Uncertainty in Multi-Robot Systems}\label{sec:uncertainty}
In this section, we outline the common forms and sources of uncertainty experienced by MRSs. 

\custompara{Outcome Uncertainty} Robot uncertainty is most commonly captured over discrete action outcomes~\cite{puterman1994mdp}, such as whether a grasp action is executed successfully.
Stochastic outcomes can occur due to robot navigation failure~\cite{ma2017multi}, battery depletion~\cite{tomy2020battery}, or stochastic features of the environment such as hazards~\cite{tihanyi2021multi}, resources~\cite{de2018preallocation}, and doors~\cite{deng2017approximate}.

\custompara{Partial Observability}
In some MRSs, robots only partially observe the environment, which prevents them from knowing each other's states.
This is often caused by limited communication and sensing capabilities, such as imperfect localisation~\cite{vanegas2017uav}, limited network range~\cite{capitan2013decentralized}, or object occlusion~\cite{hubmann2019pomdp}.
Under partial observability, robots form a \emph{belief} over the true state of the environment and other robots using possibly noisy observations obtained from sensors.

\custompara{Temporal Uncertainty}
Sources of temporal uncertainty affect the duration and start time of robot actions during execution~\cite{street2021congestion,zhang2017multirobot,boyan2001exact}.
Temporal uncertainty occurs in almost any robot environment, where action durations are affected by environmental disturbances, such as unknown obstacles or adverse weather conditions.
For example, a mobile robot's tire may slip on a carpet while navigating through an office, slowing it down.
Further, robots may have to wait for stochastic temporal processes in the environment, such as order arrival in a fulfilment centre, before beginning task execution~\cite{claes2017decentralised}.

\custompara{The Effect of Robot Interactions}
A particularly relevant driver of uncertainty in MRSs is the fact robots typically \textit{share resources}, such as space or access to a charging station, and \textit{must interact} with each other~\cite{street2022context}.
For example, when multiple mobile robots navigate in the same physical space simultaneously, they may experience \emph{congestion}, which increases uncertainty over action duration~\cite{street2021congestion}.
Alternatively, a robot manipulator may be more likely to fail a grasp if another robot is nearby, restricting its movement.
%

\section{Formal Multi-Robot Models}\label{sec:models}

In this section, we review modelling formalisms for MRSs, which we summarise in Table~\ref{table:models}.
At their foundation, each of these models consists of \emph{states}, which describe a snapshot of the MRS and environment, and \emph{transitions} between states, which define the system dynamics.

\subsection{Classical Multi-Robot Models}

\emph{Joint transition systems (JTSs)} model MRSs with deterministic dynamics~\mbox{\cite{baier2008principles,gujarathi2022mt,ulusoy2013optimality,kloetzer2011multi,kantaros2020stylus}}.
JTS states are often factored into local states for each robot, e.g. their location and battery level, and a shared set of global state features, such as whether doors in the environment are open.
%
%
JTSs are fully deterministic, and so fail to capture the stochastic dynamics of real robot environments.
\emph{Multi-agent Markov decision processes (MMDPs)} are a natural extension of JTSs to stochastic domains~\cite{boutilier1996planning}.
Similar to JTSs, MMDPs capture robots in a joint state and action space, but MMDP actions have probabilistic outcomes.
MMDPs are a common formalism for MRSs, and have been used to model drone fleets~\cite{choudhury2022scalable}, warehouse robots~\cite{claes2017decentralised}, and human-robot teams~\cite{unhelkar2020semi}.
MMDPs and JTSs assume synchronous execution, i.e. robots execute their actions in lockstep, and all actions have the same duration.
Further, the joint state and action spaces yield an exponential blow-up in the number of robots being modelled.
In practice, robot action durations are inherently continuous and uncertain, where robot interactions contribute towards this uncertainty~\cite{street2021congestion,street2022context,zhang2017multirobot,messias2013gsmdps,de2015intention}.
Thus, to accurately capture multi-robot behaviour, we require formalisms which model asynchronous multi-robot execution and uncertainty over action duration.
One approach for explicitly doing this is to use continuous-time Markov models, which we discuss later in this section.

\subsection{Avoiding the Exponential Scalability of Joint Models}

The number of MMDP or JTS states and actions increase exponentially in the number of robots~\cite{boutilier1996planning}, which makes optimal solutions for planning~\cite{kolobov2012planning}, RL~\cite{sutton2018reinforcement}, and model checking~\cite{baier2008principles} intractable.
This can be improved by making different assumptions which simplify the model.
In fact, there has been a significant research effort to identify realistic assumptions for specific multi-robot problems.
\emph{Transition-independent MMDPs (TI-MMDPs)}~\cite{scharpff2016solving} and \emph{constrained MMDPs (CMMDPs)}~\cite{de2021constrained} assume the transition dynamics of each robot are independent, but couple the MRS through rewards and shared resources, respectively.
\emph{Team MMDPs}~\cite{faruq2018simultaneous} also treat the transition dynamics independently, modelling robots sequentially in the context of simultaneous task allocation and planning problems.
Transition independence assumptions allow for weakly-coupled models that operate outside of the joint state and action space and reduce the model size, thus facilitating the use of more efficient solution methods.
%
%
However, in cases where execution-time robot interactions affect the outcome and duration of robot actions, the transition-independent models above are unable to accurately reflect the MRS.

For many multi-robot problems, robots can act independently for the majority of execution, as interactions are \textit{sparse}.
For example, two robots conducting a handover can ignore each other until they are close.
\emph{Interaction-driven Markov games (IDMGs)}~\cite{spaan2008interaction} and \emph{decentralised sparse interaction MDPs (Dec-SIMDPs)}~\cite{melo2011decentralized,melo2013heuristic} exploit this to reduce the space complexity whilst still accounting for execution-time interactions.
IDMGs and Dec-SIMDPs are equivalent, and capture an MRS using an independent MDP per robot, and a set of interaction MMDPs, which define joint MRS behaviour in interaction areas, such as near a doorway.
Though interaction MMDPs are joint models, they are significantly smaller than the full MMDP, as they are defined over only a small fraction of the full MMDP state space.
%
%
However, these models are only useful when interactions are localised to a small, fixed part of the environment.
If this does not hold, they become equivalent to the full MMDP.

%
Finally, a commonly used approach to avoid the use of joint models while still considering robot dependencies and execution-time interactions is to model the MRS as a set of single-robot models that are extended to include some knowledge of the other robots. 
In~\cite{claes2015effective,claes2017decentralised}, \emph{spatial task allocation problems (SPATAPs)} are modelled using single-robot models which aggregate the response of the other robots.
The aggregate response is represented as a distribution which predicts whether any robot is present at a given location.
This is computed by combining individual distributions over each robot's location, and allows robots to predict which tasks will be handled by other robots during planning.
A similar approach is taken in~\cite{street2021congestion}, where an MRS is modelled using single-robot \emph{time-varying Markov automata (TVMA)} which capture the probabilistic effects of congestion caused by the other robots. 
In this context, congestion is represented as a distribution over the number of robots present in each area of the environment, and distributions of navigation duration under the presence of a specific number of robots are obtained from real-world multi-robot navigation data.
To solve multi-robot planning problems,~\cite{zhang2017multirobot} augment single-robot models with a cost function which captures the effects of robot interactions.
This cost function is then adjusted iteratively during planning to encourage robot collaboration.

\subsection{Partially Observable Multi-Robot Models}

Partially observable MDPs (POMDPs) are widely used to model partially observable problems, where robots make  observations which update their belief over their current state~\cite{kaelbling1998planning}.
\emph{Decentralised POMDPs (Dec-POMDPs)} extend POMDPs to multi-robot settings~\cite{bernstein2002complexity}, where each robot has its own set of local observations.
Dec-POMDPs have been used for warehouse robotics~\cite{amato2015planning}, cooperative package delivery~\cite{omidshafiei2017decentralized}, and teams of unmanned aerial vehicles~\cite{floriano2019planning}.
If the combined local observations of each robot uniquely identify the joint state, Dec-POMDPs are reduced to Dec-MDPs, which are easier to solve~\cite{bernstein2002complexity}.
However,  these are still joint models, and optimal solvers for both Dec-POMDPs and Dec-MDPs have even higher time complexity than MMDP solvers~\cite{bernstein2002complexity}.
%
To reduce the space complexity related to the joint modelling in Dec-POMDPs,~\cite{oliehoek2012influence,oliehoek2021sufficient} consider decoupling them into local POMDPs for each robot.
For each of these local POMDPs, they compute a distribution which captures how external state factors influence its local state.
These external state factors include the states of the other robots.
This is then used to marginalise out the external state factors to construct single-robot POMDPs.
This \textit{influence-based abstraction} produces smaller models.
However, computing influence distributions is intractable in general~\cite{oliehoek2021sufficient}.

%
Another class of relevant POMDP-based models are \emph{macro action Dec-POMDPs (MacDec-POMDPs)}~\cite{amato2019modeling} and \emph{decentralised partially observable semi-MDPs (Dec-POSMDPs)}~\cite{omidshafiei2017decentralized}, which  consider \textit{macro actions} which execute a series of primitive low-level actions, such as moving one grid cell forward.
This hierarchical paradigm is based on the options framework~\cite{sutton1999between} for MDPs and has two main benefits.
First, it reduces model size by leveraging existing behaviour such as navigation, and modelling behaviour at the macro action level, rather than each time step.
Second, the use of temporally extended actions seamlessly enables asynchronous action execution.
Each MacDec-POMDP and Dec-POSMDP has an underlying Dec-POMDP which captures the low-level actions that form the macro actions.
For MacDec-POMDPs, the underlying Dec-POMDP and the policies for each macro action are assumed to be known~\cite{amato2018decision}.
MacDec-POMDP policies can then be evaluated by unrolling the macro actions on the low-level Dec-POMDP.
Unlike MacDec-POMDPs, Dec-POSMDPs capture macro actions using distributions over their completion time, where Dec-POSMDP policies can be evaluated through simulation.

\subsection{Continuous-Time Multi-Robot Models}

Several models have been proposed to take into account uncertainty over action duration in the context of MRSs which are evolving asynchronously.
These make use of continuous-time distributions which capture the stochasticity in robot action durations.
\emph{Continuous-time MDPs (CTMDPs)} extend MDPs to include durative transitions represented as exponential delays~\cite{guo2009continuous}, and have been used to model multi-robot data collection problems~\cite{yin2011continuous}.
To model asynchronous multi-robot execution, CTMDPs can be defined over a joint state and action space, similar to MMDPs.
Thus, as with MMDPs, they scale exponentially in the number of robots.
To mitigate this,~\cite{yin2011continuous} constructs single-robot CTMDPs assuming transition independence, similar to~\cite{scharpff2016solving,de2021constrained}.
The duration of each action in a CTMDP is modelled with a single exponential distribution.
This is a convenience which allows for simpler solution approaches which exploit the memoryless property of the exponential distribution, but limits the accuracy with which we can capture robot action durations.

Many multi-robot models can capture \emph{heterogeneous} MRSs (see Table~\ref{table:models}), where robots have different capabilities and resource usage etc.
This is often achieved using local action spaces or reward functions for each robot.
\emph{Generalised stochastic Petri nets (GSPNs)}~\cite{balbo2007introduction} are a modelling formalism for \emph{homogeneous} MRSs, i.e. the robots are identical, where robots are represented \emph{anonymously} as tokens.
Further, as in CTMDPs, durations are restricted to exponentials.
GSPNs remain exponential in the team size, but robot anonymity provides a practical reduction in the number of states.
GSPNs have been used to model teams of football robots~\cite{Costelha2012}, autonomous haulers~\cite{mansouri2019multi}, and monitoring robots~\cite{azevedo2020iros}.
%
\emph{Generalised semi-MDPs (GSMDPs)} can capture concurrent execution and stochastic durations, and have been applied to MRSs in~\cite{younes2004solving,messias2013gsmdps}, but are complex to define and hard to solve, as GSMDPs allow for arbitrary duration distributions.
\emph{Multi-robot Markov automata (MRMA)}~\cite{street2022context} also allow for arbitrary duration distributions to capture asynchronous multi-robot execution in continuous time.
Markov automata (MA) extend MDPs and CTMDPs by explicitly separating instantaneous robot action choice and the duration of robot actions~\cite{eisentraut2010probabilistic}.
MRMA are joint models, where robot action durations are represented as phase-type distributions (PTDs), which are sequences of exponentials capable of capturing any nonnegative distribution to an arbitrary level of precision~\cite{buchholz2014input}.
In an MRMA, there is a different duration distribution for each spatiotemporal situation an action may be executed under, referred to as the \emph{context}, which captures the effects of robot interactions on action execution.
By separating robot decision making from action duration, robot interactions can be detected at the instant an action is triggered by analysing the joint MRMA state.
MRMA are connected to other continuous-time multi-robot models.
First, GSPN semantics can be described with an MA~\cite{eisentraut2013petrinets}.
Second, a standard solution for GSMDPs involves converting all duration distributions into PTDs~\cite{buchholz2014input}, which produces a model similar to an MRMA~\cite{younes2004solving}.
However, MRMA are simpler to define, and can be solved directly~\cite{hatefi2012model}, as all durations are exponentials/PTDs by definition.

\begin{sidewaystable}
\centering
\footnotesize
\caption{A summary of multi-robot modelling formalisms.}
\begin{tabular}{|l|c|c|c|c|c|c|c|}
\hline
\textbf{Model} & \textbf{\makecell{Stochastic \\ Outcomes}} & \textbf{\makecell{Partial \\ Observability}} & \textbf{\makecell{Temporal \\ Uncertainty}} & \textbf{\makecell{Continuous \\ Time}} & \textbf{\makecell{Transition \\ Independence}} & \textbf{\makecell{Asynchronous \\ Execution}} & \textbf{\makecell{Allows for \\ Heterogeneous \\ Teams}} \\ \hline
JTS~\cite{baier2008principles} & \xmark & \xmark & \xmark & \xmark& \xmark& \xmark& \cmark\\ \hline
MMDP~\cite{boutilier1996planning} & \cmark & \xmark & \xmark & \xmark& \xmark& \xmark& \cmark\\ \hline
TI-MMDP~\cite{scharpff2016solving} & \cmark & \xmark & \xmark & \xmark& \cmark& \xmark& \cmark \\ \hline
CMMDP~\cite{de2021constrained} & \cmark & \xmark & \xmark & \xmark& \cmark& \xmark& \cmark \\ \hline
Team MMDP~\cite{faruq2018simultaneous} & \cmark & \xmark & \xmark & \xmark& \xmark& \xmark& \cmark \\ \hline
Dec-SIMDP/IDMG~\cite{melo2011decentralized,spaan2008interaction} & \cmark & \xmark & \xmark & \xmark& \xmark& \xmark& \cmark \\ \hline
SPATAP Model~\cite{claes2015effective} & \cmark & \xmark& \xmark& \xmark& \cmark& \xmark& \cmark\\\hline
TVMA per Robot~\cite{street2021congestion} & \cmark & \xmark & \cmark & \xmark & \xmark & \cmark & \cmark \\\hline
Dec-POMDP~\cite{bernstein2002complexity} & \cmark & \cmark & \xmark & \xmark& \xmark& \xmark& \cmark \\ \hline
MacDec-POMDP~\cite{amato2019modeling} & \cmark & \cmark & \cmark & \xmark& \xmark& \cmark& \cmark \\ \hline
Dec-POSMDP~\cite{omidshafiei2017decentralized} & \cmark & \cmark & \cmark & \xmark& \xmark& \cmark& \cmark \\ \hline
CTMDP~\cite{guo2009continuous} & \cmark & \xmark & \cmark & \cmark& \xmark& \cmark& \cmark \\ \hline
GSPN~\cite{balbo2007introduction} & \cmark & \xmark & \cmark & \cmark& \cmark& \cmark& \xmark \\ \hline
GSMDP~\cite{younes2004solving} & \cmark & \xmark & \cmark & \cmark & \xmark& \cmark& \cmark\\ \hline
MRMA~\cite{street2022context} & \cmark & \xmark & \cmark & \cmark & \xmark& \cmark& \cmark\\ \hline
\end{tabular}
\label{table:models}
\end{sidewaystable}

\section{Model Applications}\label{sec:apps}

\begin{table}
\setlength{\tabcolsep}{5pt}
\centering
\footnotesize
\caption{Applications of the models in Table~\ref{table:models} for multi-robot/multi-agent problems.}
\begin{tabular}{|l|c|c|c|c|}
\hline
\textbf{Model} & \textbf{Planning} & \textbf{\makecell{Reinforcement \\ Learning}} & \textbf{\makecell{Model \\ Checking}} & \textbf{\makecell{Simulation}}\\ \hline
JTS~\cite{baier2008principles} & \cite{gujarathi2022mt,ulusoy2013optimality,kloetzer2011multi,kantaros2020stylus} & - & \cite{gujarathi2022mt,ulusoy2013optimality,kloetzer2011multi,kantaros2020stylus} & -\\ \hline
MMDP~\cite{boutilier1996planning} & \cite{boutilier1996planning,choudhury2022scalable,NIKOU201710766,wongpiromsarn2013incremental} & \cite{melcer2022shield,yang2020multi} & \cite{NIKOU201710766,wongpiromsarn2013incremental} & -\\ \hline
TI-MMDP~\cite{scharpff2016solving} & \cite{scharpff2016solving} & - & - & -\\ \hline
CMMDP~\cite{de2021constrained} & \cite{dolgov2006resource,de2017bounding,de2020risk,agrawal2016scalable,de2018preallocation,2023AAAI_gautier,2023AAMAS_gautier} & \cite{lu2021decentralized,he2022robust} & - & -\\ \hline
Team MMDP~\cite{faruq2018simultaneous} & \cite{faruq2018simultaneous} & - & \cite{faruq2018simultaneous} & - \\ \hline
Dec-SIMDP/IDMG~\cite{melo2011decentralized,spaan2008interaction} & \cite{melo2011decentralized,melo2013heuristic,spaan2008interaction} & \cite{ganguly2022decentralized,kujirai2019greedy} & - & -\\ \hline
SPATAP Model~\cite{claes2015effective} & \cite{claes2015effective,claes2017decentralised} & - & - & - \\\hline
TVMA per Robot~\cite{street2021congestion} & \cite{street2021congestion} & - & - & -\\\hline
Dec-POMDP~\cite{bernstein2002complexity} & \cite{floriano2019planning,omidshafiei2017scalable,floriano2021hybrid,lauri2020multi} & \cite{rashid2018qmix,xiao2021local,jiang2021multi,lyu2020likelihood,omidshafiei2019learning,peng2021facmac,pan2021regularized,gupta2021uneven,willemsen2021mambpo} & - & -\\ \hline
MacDec-POMDP~\cite{amato2019modeling} & \cite{amato2014planning,amato2015planning,amato2016policy,amato2019modeling,hoang2018near} & \cite{xiao2020macro,xiao2020learning,liu2017learning,xiao2022asynchronous} & - & - \\ \hline
Dec-POSMDP~\cite{omidshafiei2017decentralized} & \cite{omidshafiei2017decentralized,omidshafiei2017scalable,omidshafiei2017semantic} & - & - & - \\ \hline
CTMDP~\cite{guo2009continuous} & \cite{yin2011continuous,jia2015continuous} & - & - & -\\ \hline
GSPN~\cite{balbo2007introduction} & \cite{mansouri2019multi,azevedo2020iros,azevedo2021petri,chen2011assembly} & - & - & \cite{Costelha2012,chen2011assembly}\\ \hline
GSMDP~\cite{younes2004solving} & \cite{messias2013gsmdps} & - & - & - \\ \hline
MRMA~\cite{street2022context} & - & - & - & \cite{street2022context}\\ \hline
\end{tabular}
\label{table:apps}
\end{table}

In this section, we discuss how the multi-robot models in Table~\ref{table:models} have been solved and analysed for multi-robot planning, RL, model checking, and simulation.
We  summarise this discussion in Table~\ref{table:apps}.
Note that in Table~\ref{table:apps} we do not list foundational works which apply to more general models, such as heuristic search approaches for MDPs which can be applied to MMDPs~\cite{kolobov2012planning}, or MA model checking techniques which can be applied to MRMA~\cite{hatefi2012model}.

\subsection{Planning}

Multi-robot planning techniques synthesise robot behaviour given a formal model of the system.
%
%
Many multi-robot models can be solved with standard techniques.
MMDPs can be solved exactly using MDP solvers such as value or policy iteration~\cite{bellman1966dynamic,howard1960dynamic}.
However, these methods solve for all states, making them intractable for joint multi-robot models.
Heuristic and sampling-based methods such as labelled real-time dynamic programming~\cite{bonet2003labeled} or Monte-Carlo tree search~\cite{kocsis2006bandit} improve upon the limited scalability of exact solvers by restricting search to promising areas of the state space.
Despite reducing the explored states, heuristic algorithms are slow to converge on large models, but often provide anytime behaviour such that valid solutions are synthesised quickly, and improved with time.
The poor scalability of MMDP planning motivates planning on simplified models.
For TI-MMDPs~\cite{scharpff2016solving}, transition independence allows for compact representations of reward dependencies in conditional return graphs, which admits efficient solutions.
For Dec-SIMDPs and IDMGs, the single-robot MDPs and interaction MMDPs can be solved separately using standard solvers such as value iteration~\cite{spaan2008interaction}.
Similarly, the SPATAP models in~\cite{claes2015effective} are single-robot MDPs which capture the effects of the other robots, and can be solved separately.
CMMDP approaches typically exploit the fact that only the resource constraint couples the agents to scale to larger problems.
Planning for CMMDPs has considered a range of constraints over resource consumption, such as bounding its worst-case~\cite{dolgov2006resource}, considering a chance-constraint~\cite{de2017bounding,2023AAAI_gautier}, and bounding its conditional value at risk~\cite{2023AAMAS_gautier}.

MMDPs can be solved tractably if they are sufficiently small.
Therefore, in~\cite{NIKOU201710766} robots are grouped into clusters based on robot dependencies, and each cluster is solved as a separate MMDP.
Similarly, in~\cite{wongpiromsarn2013incremental} robots are incrementally added to an MMDP to control scalability.

Recent work~\cite{choudhury2022scalable} has begun to address the poor scalability of MMDP planning.
There, an anytime planner for MMDPs based on Monte Carlo tree search is presented, where robot dependencies are exploited to decompose the value function into a set of factors from which the optimal joint action can be computed.
This approach scales to previously intractable problems.

Solution methods for continuous-time multi-robot models differ depending on the objective.
To solve CTMDPs for time-abstract objectives, such as expected untimed reward, MDP solvers are applied to an embedded time-abstract MDP.
For timed objectives, MDP solvers are instead applied to a uniformised MDP, where each state has the same expected sojourn time~\cite{kakumanu1977relation,younes2004solving,butkova2017long}.
Similarly, GSPNs can be converted to an MDP~\cite{mansouri2019multi} or an MA~\cite{azevedo2020iros} dependent on the objective and solved with standard techniques.
For MRMA, we can plan using MA solution methods~\cite{hatefi2012model}.

Dec-POMDPs can be solved centrally to synthesise local policies for decentralised execution, which map from local action-observation histories to actions~\cite{floriano2019planning,omidshafiei2017scalable,floriano2021hybrid,lauri2020multi}.
With this, local Dec-POMDP policies are robust to communication limitations and unreliable sensors.
Dec-POMDP solutions can be adapted to MacDec-POMDPs and Dec-POSMDPs to synthesise policies over macro actions.
In~\cite{amato2014planning}, the space of macro-action policies is searched exhaustively, where efficient simulators improve the scalability of policy evaluation~\cite{amato2015planning}.
This approach scales poorly, which is addressed in~\cite{amato2016policy}, where a heuristic search method optimises finite state controllers for each robot.
However, MacDec-POMDP and Dec-POSMDP solutions have not been shown to scale beyond teams of around four robots~\cite{amato2019modeling,omidshafiei2017decentralized}.

\subsection{Reinforcement Learning (RL)}

An alternative approach to policy synthesis is RL~\cite{sutton2018reinforcement}.
Planners synthesise behaviour using a model of the system, whereas RL approaches learn behaviour using data sampled from the environment~\cite{kolobov2012planning,sutton2018reinforcement}.
Multi-robot RL problems are formulated assuming an underlying multi-robot model which is unknown prior to training.
Fully observable, centralised problems can be formulated as an MMDP~\cite{melcer2022shield,yang2020multi} and solved using standard RL techniques such as deep Q-learning~\cite{mnih2015human}.
However, these techniques do not scale to multi-robot problems due to the exponential increase in the state and action space~\cite{yang2020multi,rashid2018qmix}.
In many settings, decentralised policies are required due to limited communication or partial observability~\cite{rashid2018qmix,xiao2021local}.
Here, multi-robot RL can be formulated as a Dec-POMDP and solved  under the paradigm of centralised training with decentralised execution~\cite{oliehoek2008optimal}, which allows additional state information not available during execution to be used during training, such as the joint state.
One example of this paradigm is QMix~\cite{rashid2018qmix}, which uses a mixing network to estimate the joint Q value from single-robot Q values.
RL techniques for Dec-POMDPs are still slow to converge however, and so MacDec-POMDPs can be used to exploit existing behaviours and improve the efficiency of learning~\cite{xiao2020macro,xiao2020learning,liu2017learning,xiao2022asynchronous}.

\subsection{Model Checking}

Model checking techniques evaluate the behaviour induced by robot policies by systematically checking if a property is satisfied in a formal robot model~\cite{baier2008principles}.
Properties are often specified with temporal logics such as linear temporal logic (LTL) or continuous stochastic logic (CSL).
Similar to planning, many of the multi-robot models in Table~\ref{table:models} can be verified using techniques for more general models.
For example, LTL formulae can be verified on JTSs and MMDPs using techniques for transition systems and MDPs~\cite{baier2008principles}.
However, exact LTL model checking approaches compute a product of the model and an automaton that captures the LTL formula, which significantly increases the state space, making them unsuitable for multi-robot problems.
MRMA can be model checked against CSL formulae using model checking techniques for MA~\cite{hatefi2012model}.
This also applies to GSPNs, which can be represented as an MA with identical semantics~\cite{eisentraut2013petrinets}.
Similar CSL model checking techniques are available for CTMDPs~\cite{buchholz2011model}.

Model checking and planning are often combined to synthesise guaranteed multi-robot behaviour.
For LTL specifications, we can plan over a joint product automaton, however this quickly becomes intractable.
To overcome this,~\cite{faruq2018simultaneous} concatenate single robot product automata through switch transitions in a team MMDP to reduce the state space.
For MMDPs, in~\cite{wongpiromsarn2013incremental} robots are added incrementally to a product automaton until the full problem is solved, or a fixed computational budget is exceeded.
Alternatively, in~\cite{kantaros2020stylus} the product automaton is explored incrementally through sampling for MRSs modelled as a JTS.
Combined planning and model checking techniques have been used for multi-robot data gathering~\cite{gujarathi2022mt,ulusoy2013optimality}, monitoring~\cite{kloetzer2011multi}, and mobility-on-demand~\cite{wongpiromsarn2013incremental}.

Statistical model checking (SMC) techniques evaluate properties by sampling through a model given a set of robot policies, which avoids enumerating the state space~\cite{legay2010statistical}, and bridges the gap between model checking and simulation techniques, which we discuss later in this section.
In~\cite{herd2015quantitative}, SMC is used to evaluate quantitative properties of an MRS.
SMC techniques can be applied to many of the models in Table~\ref{table:models}.
For example, we can use SMC techniques for MA~\cite{butkova2021modest} to evaluate bounded or unbounded properties on an MRMA.
A drawback of SMC is a possible failure to explore states reached with low probability, which can render SMC unsuitable for safety critical systems~\cite{butkova2021modest}.

\subsection{Simulation}

Simulators evaluate multi-robot behaviour by executing a set of robot policies in an abstracted environment model.
Using formal multi-robot models, we can create a discrete-event simulator (DES) by sampling stochastic outcomes and durations, and resolving non-determinism using robot policies.
DESs mitigate the complexity of physics-based simulators such as Gazebo~\cite{koenig2004design} by abstracting away low-level robot dynamics~\cite{bakker2013rams}, allowing simulations to run magnitudes faster than real time.
GSPNs, or variants thereof, have been used to simulate teams of football robots~\cite{Costelha2012} and human-robot manufacturing teams~\cite{chen2011assembly}, respectively.
In~\cite{street2022context}, a DES called CAMAS (context-aware multi-agent simulator) samples through an MRMA to evaluate task-level metrics of multi-robot performance under the effects of robot interactions, such as the time to complete a set of tasks.

\section{Conclusions}\label{sec:conclusions}

In this paper, we reviewed modelling approaches for capturing the task-level behaviour of MRSs.
We focused on stochastic models of multi-robot execution, and introduced the different types of uncertainty encountered by MRSs.
Further, we discussed how these models have been used for multi-robot planning, RL, model checking, and simulation.
Recent research has focused on constructing models which accurately capture the effects of uncertainty and robot interactions, or constructing models small enough to be solved efficiently.
These two objectives are opposing, as to accurately capture multi-robot execution, we often require joint models which are frequently intractable to solve or analyse.
Therefore, future research should focus on developing smaller multi-robot models which still accurately capture uncertainty and robot interactions.
This may be achieved by identifying realistic assumptions over the sources of uncertainty and robot interactions, such as interactions only occurring in small portions of the state space.
Exploiting these assumptions allows for smaller models which can be solved efficiently without sacrificing model accuracy.
An alternative avenue for research is to exploit the structure of multi-robot problems, such as factored state spaces and dependencies between robots, to develop scalable solution methods for multi-robot models.

\backmatter

\bmhead{Acknowledgments}
Charlie Street and Masoumeh Mansouri are UK participants in Horizon Europe Project CONVINCE, and supported by UKRI grant number 10042096.
Bruno Lacerda is supported by the EPSRC Programme Grant `From Sensing to Collaboration' (EP/V000748/1).

\section*{Declarations}

\bmhead{Conflict of Interest} The authors declare no competing interests.
\bmhead{Human and Animal Rights and Informed Consent} This article does not
contain any studies with human or animal subjects performed by any
of the authors.

\section*{Important References}
In this section, we highlight recently published papers of particular interest.

\subsection*{Important}
\begin{itemize}
    \item \cite{de2021constrained}: This paper provides a comprehensive taxonomy of CMMDP problems and solutions, and is an effective starting point for new researchers in the area.
    \item \cite{oliehoek2021sufficient}: This paper formalises influence-based abstraction for decomposing Dec-POMDPs into single-robot models without sacrificing task performance.
\end{itemize}
\subsection*{Very Important}
\begin{itemize}
\item \cite{street2022context}: This paper introduces MRMA, the first formulation to explicitly capture the effects of execution-time robot interactions on action duration, as well as CAMAS, which samples through an MRMA to evaluate task-level metrics of multi-robot performance.
\item \cite{amato2019modeling}: This paper presents MacDec-POMDPs, and extends three Dec-POMDP solvers to handle macro actions. 
Further, this paper demonstrates how planning with macro actions can scale to previously intractable Dec-POMDP problems.
\item \cite{rashid2018qmix}: This paper presents QMix, a state of the art reinforcement learning approach for multi-robot problems formulated as a Dec-POMDP.
\item \cite{choudhury2022scalable}: This paper presents a state of the art anytime planner for MMDPs based on Monte Carlo tree search which can solve previously intractable problems.
\end{itemize}


\bibliography{references}


\end{document}